\documentclass[conference]{IEEEtran}
\pdfoutput=1

\usepackage{cite}
\usepackage{amsmath,amssymb,amsfonts}
\usepackage{algorithmic}
\usepackage{graphicx}
\usepackage{textcomp}
\usepackage{xcolor}

\graphicspath{{./}{./figs/}}

\usepackage{hyperref}

\begin{document}

\title{A Topology-Aware Graph Convolutional Network for Human Pose Similarity and Action Quality Assessment}

\author{
\IEEEauthorblockN{Minmin Zeng}
\IEEEauthorblockA{Shenzhen TSAF Tech. Co. Ltd.\\
\texttt{minmin.zeng@tsaftech.com}}
}

\maketitle

\begin{abstract}
Action Quality Assessment (AQA) is a challenging task that requires a fine-grained understanding of human motion. Existing methods for pose similarity assessment often rely on direct coordinate comparisons or generic feature extractors, which may fail to capture the inherent topological structure of the human skeleton. This paper introduces a novel framework for pose similarity assessment based on a Graph Convolutional Network (GCN). We model the human skeleton as a graph structure, enabling the GCN to learn topology-aware feature embeddings that are robust to variations in position and scale. The network is trained using a Contrastive Regression paradigm to effectively discriminate between poses of varying similarity. Our pipeline consists of a top-down human pose estimator (YOLOv5 and HRNet) to extract 2D keypoints, which are then fed into our proposed GCN-based feature extractor. We conduct rigorous evaluations on standard AQA benchmarks, including AQA-7 and FineDiving. The results demonstrate that our method achieves comparable or superior performance to state-of-the-art approaches, validating the effectiveness of learning topological features for pose similarity assessment.
\end{abstract}


\section{Introduction}

Action Quality Assessment (AQA), an important subfield of computer vision, has garnered significant attention in recent years. Its core objective is to automatically quantify the execution quality of human actions \cite{aqa_survey_2024}. AQA technology exhibits immense potential across numerous domains. For instance, in sports analytics, it can provide objective scores for athletes in disciplines like diving and gymnastics. In medical rehabilitation, it can monitor whether a patient's therapeutic exercises are performed correctly. In professional training, such as for firefighters or surgeons, it can offer immediate and accurate feedback, thereby enhancing training efficiency and safety \cite{aqa_comprehensive_2024}.

In AQA tasks, human pose serves as the fundamental basis for analyzing action quality. By abstracting the human body into a skeleton composed of keypoints (joints) and bones (limbs), it is possible to effectively mitigate the influence of irrelevant variables such as clothing, background, and lighting, thereby focusing on the action itself \cite{action_recognition_review}. Consequently, evaluating the quality of an action largely reduces to precisely comparing the performer's pose against a standard or reference pose \cite{beat_alignment}. A robust and accurate method for quantifying pose similarity is therefore critical for building high-performance AQA systems.

However, existing methods for calculating pose similarity have notable limitations. Early approaches based on direct pixel-wise image comparison are highly susceptible to appearance variations and lack the robustness required for practical applications. With the advancement of deep learning-based pose estimation algorithms, methods based on keypoint coordinates have become mainstream. These methods typically flatten the 2D or 3D coordinates of all keypoints into a single long vector and then compute a distance metric (e.g., Euclidean or cosine distance) between these vectors \cite{pose_similarity_ml}. While this approach is viable to an extent, it disregards a crucial piece of information: the intrinsic topological structure of the human skeleton. In this vectorized representation, the relationship between a wrist and an elbow is treated identically to the relationship between a wrist and an ankle, failing to reflect the inherent constraints of the human kinematic chain.

To address these shortcomings, this paper posits a central hypothesis: explicitly modeling the human skeleton as a graph and leveraging a Graph Convolutional Network (GCN) for feature extraction will enable the learning of more powerful and semantically meaningful pose representations, thereby achieving more accurate pose similarity assessment. Based on this premise, we propose a novel, end-to-end framework for pose similarity evaluation. The framework first employs a state-of-the-art Convolutional Neural Network (CNN) for human pose estimation to precisely locate keypoints. Subsequently, these keypoints and their natural connections are used to construct a skeletal graph. Finally, a specially designed GCN model learns discriminative, topology-aware feature embeddings from this graph for the final similarity computation.

The primary contribution of this work is the explicit application of GCNs to learn discriminative embeddings for individual static poses to solve a fine-grained quality assessment problem. This stands in contrast to the large body of existing research that uses GCNs for temporal action recognition, which primarily focuses on classifying entire sequences of motion. Our method provides a new, structure-aware solution to a core sub-problem within AQA: static pose evaluation.

\section{Related Work}

\subsection{Human Pose Estimation}

Human Pose Estimation aims to localize the anatomical keypoints of the human body from images or videos, forming a foundational step for understanding human behavior \cite{pose_survey}. Modern deep learning-based HPE methods are primarily categorized into two paradigms: bottom-up and top-down \cite{topdown_survey}. Bottom-up approaches first detect all keypoints in an image and then group them into individual human instances. In contrast, top-down approaches first use an object detector to locate each person within a bounding box and then estimate the pose for each person independently. While computationally more intensive, this strategy typically achieves higher accuracy.

Among the many pose estimation architectures, the High-Resolution Network (HRNet) has demonstrated outstanding performance \cite{hrnet_sensor,hrnet_spie}. Unlike conventional networks that downsample to extract semantic information and then upsample to recover resolution, HRNet maintains high-resolution feature maps throughout the process and repeatedly fuses information across multi-resolution branches. This design enables HRNet to generate spatially precise keypoint heatmaps, making it highly suitable for AQA tasks that demand high localization accuracy.

\subsection{Skeleton-based Action Analysis}

Analyzing actions using human skeleton data offers significant advantages. Skeleton data is a highly abstract representation that is naturally robust to variations in lighting, complex backgrounds, and clothing \cite{action_recognition_review}. In the field of action analysis, it is important to distinguish between Action Recognition and Action Quality Assessment (AQA). Action recognition aims to classify an action sequence into a predefined category. Large-scale datasets like NTU RGB+D were created to advance research in action recognition \cite{ntu_rgbd,ntu_wiki}. AQA, on the other hand, is a more complex task that evaluates the execution quality of an action. This paper focuses on the AQA domain, particularly the core problem of pose similarity assessment.

\subsection{Graph Convolutional Networks for Human Skeletons}

The emergence of Graph Convolutional Networks (GCNs) has provided a powerful tool for processing non-Euclidean data structures like graphs \cite{stgae}. Given that the human skeleton is naturally a graph structure (joints as nodes, bones as edges), GCNs were quickly applied to skeleton-based action analysis. A pioneering work in this area is the Spatio-Temporal Graph Convolutional Network (ST-GCN) \cite{stgcn_original,part_gcn}. ST-GCN constructs a spatio-temporal graph that includes both spatial dimensions and temporal dimensions to capture the dynamic evolution of an action sequence \cite{stgcn_traffic,feedback_gcn}.

However, it is crucial to emphasize that while our method is inspired by these works, its application is fundamentally different. The core of models like ST-GCN is to process temporal sequences for classification \cite{stgcn_emergent}. Our method, in contrast, focuses on learning a discriminative feature embedding for a single, static human pose to be used for quantifying similarity between poses.

\subsection{Action Quality Assessment}

AQA research aims to develop algorithms that can mimic the scoring process of expert judges \cite{aqa_survey_2024}. Existing methods can be broadly classified into regression-based methods and ranking/contrastive methods. Among these, the Contrastive Regression (CoRe) framework \cite{core_github,core_iccv} has become a representative work. The core idea of CoRe is to learn the relative score difference between pairs of videos rather than absolute scores. Recently, Transformer-based models like the Temporal Parsing Transformer (TPT) \cite{tpt_eccv,tpt_researchgate} have further improved AQA performance by decomposing actions into semantically meaningful stages.

\section{GCN-based Pose Similarity Network}

\subsection{Overall Architecture}

The GCN-PSN employs a weight-sharing Siamese Network architecture. The architecture consists of four core stages: Human Pose Estimation, Topology-Aware Feature Extraction, Network Training via Contrastive Regression, and Pose Similarity Scoring. During inference, two input images are passed through identical pipelines to generate two 50-dimensional feature vectors. The cosine distance between these vectors serves as the raw measure of pose similarity.

\begin{figure}[t]
  \centering
  \includegraphics[width=\columnwidth]{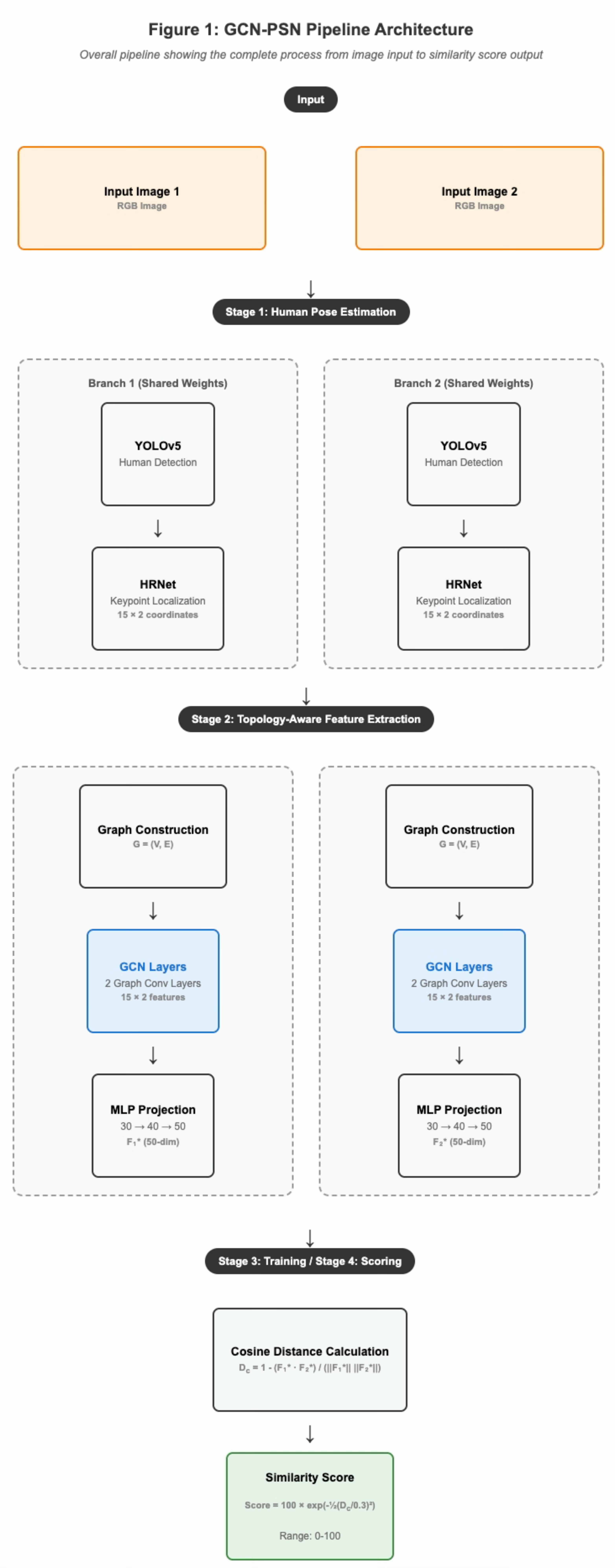}
  \caption{Overall pipeline of the GCN-PSN. This flowchart illustrates the complete process from image input to the final similarity score, with the core being the use of a GCN to extract topology-aware keypoint features.}
  \label{fig:pipeline}
\end{figure}

\subsection{Human Pose Estimation}

The goal of this stage is to accurately extract the 2D coordinates of human skeletal keypoints from the raw RGB image. We adopt a standard top-down approach.

\textbf{Human Detection:} We use a fine-tuned YOLOv5 model to detect humans in the image. The standard YOLOv5 is modified from an 80-class task to a single-class task, detecting only the ``person'' category. For each detected person, the model outputs the bounding box coordinates $(x_{\text{min}}, y_{\text{min}}, x_{\text{max}}, y_{\text{max}})$.

\begin{figure*}[t]
  \centering
  \includegraphics[width=\textwidth]{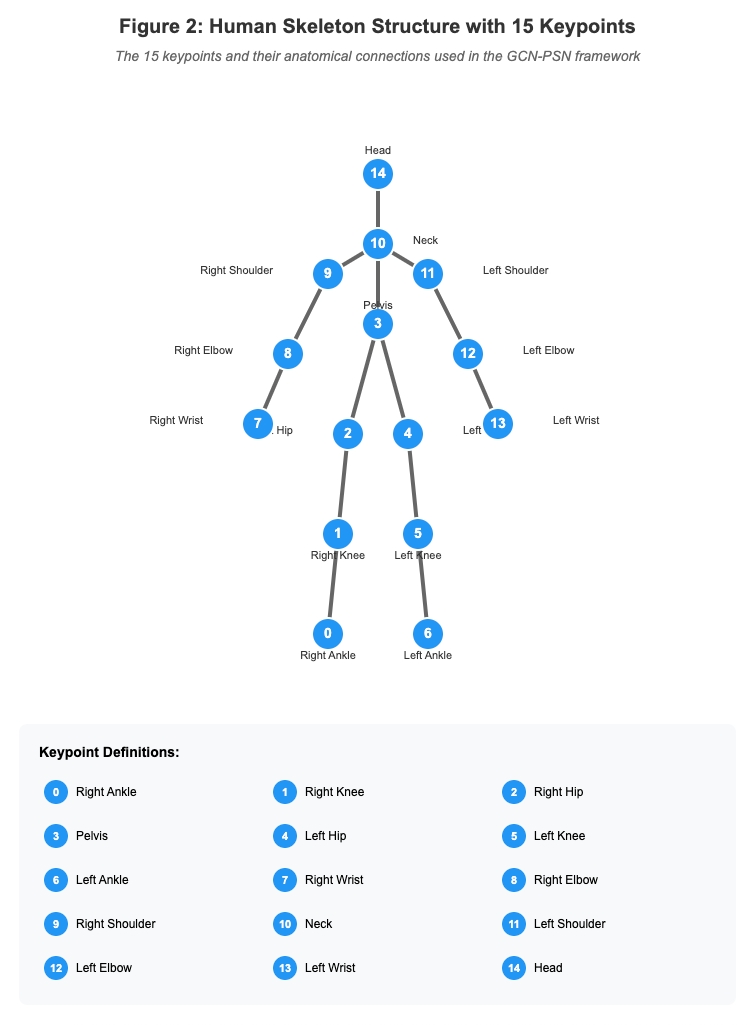}
  \caption{The 15 human skeleton keypoints and their connections used in this paper. The numbering is: 0-Right Ankle, 1-Right Knee, 2-Right Hip, 3-Pelvis, 4-Left Hip, 5-Left Knee, 6-Left Ankle, 7-Right Wrist, 8-Right Elbow, 9-Right Shoulder, 10-Neck, 11-Left Shoulder, 12-Left Elbow, 13-Left Wrist, 14-Head.}
  \label{fig:skeleton}
\end{figure*}

\textbf{Keypoint Localization:} After obtaining the human bounding box, we crop the image to that region and feed it into HRNet for keypoint localization. We select 15 keypoints that are primarily involved in motion and number them from 0 to 14, as shown in Fig.~\ref{fig:skeleton}.

\subsection{Topology-Aware Feature Extraction}

This is the core innovation of our framework. This stage transforms the discrete keypoint coordinates into a robust feature vector that captures the structural information of the human body.

\textbf{Pose Graph Construction:} We formally define the 15 extracted keypoints as an undirected graph $G=(V,E)$.
\begin{itemize}
\item \textbf{Node Set} $(V)$: The node set $V = \{v_i \mid i=0, \ldots, 14\}$ represents the 15 human keypoints. The initial attribute of each node $v_i$ is its 2D coordinate $(x_i, y_i)$.
\item \textbf{Edge Set} $(E)$: The edge set $E$ represents the 14 bones connecting adjacent joints.
\item \textbf{Adjacency Matrix} $(C)$: We construct a $15 \times 15$ adjacency matrix $C$. If an edge exists between node $i$ and $j$, or if $i=j$ (self-loop), then $C_{ij}=1$; otherwise, $C_{ij}=0$.
\end{itemize}

\textbf{Node Feature Initialization and Normalization:} To make the pose representation invariant to position and size, we normalize the coordinates. For all 15 keypoints, we find $x_{\text{min}}, x_{\text{max}}, y_{\text{min}}, y_{\text{max}}$. Then, coordinates $(x_i, y_i)$ are normalized to $(x_i^*, y_i^*)$:
\begin{equation}
x_i^* = \frac{x_i - x_{\text{min}}}{x_{\text{max}} - x_{\text{min}}}
\end{equation}
\begin{equation}
y_i^* = \frac{y_i - y_{\text{min}}}{y_{\text{max}} - y_{\text{min}}}
\end{equation}
The normalized coordinates form an initial $15\times 2$ feature matrix $X$.

\textbf{Graph Convolutional Network:} We employ a GCN model with two graph convolutional layers. The propagation rule for each layer is:
\begin{equation}
H^{(l+1)} = \sigma(\hat{D}^{-\frac{1}{2}}\hat{C}\hat{D}^{-\frac{1}{2}}H^{(l)}W^{(l)})
\end{equation}
where $H^{(l)}$ is the node feature matrix at layer $l$ ($H^{(0)}=X$), $\hat{C}=C+I$ is the adjacency matrix with self-loops, $I$ is the identity matrix, $\hat{D}$ is the degree matrix of $\hat{C}$ with $\hat{D}_{ii}=\sum_j \hat{C}_{ij}$, $W^{(l)}$ is the trainable weight matrix, and $\sigma$ is the ReLU activation function.

\textbf{Feature Flattening and Projection:} After GCN processing, we flatten the $15\times 2$ feature matrix into a 30-dimensional vector $F$. We then feed $F$ into a two-layer MLP with structure: Input (30-dim) $\rightarrow$ Hidden 1 (40-dim) $\rightarrow$ Hidden 2 (50-dim) $\rightarrow$ Output (50-dim), resulting in a final 50-dimensional pose embedding vector $F^*$.

\subsection{Training via Contrastive Regression}

We adopt a contrastive regression training strategy using a Siamese network architecture. In each training step, we randomly select a pair of pose samples and obtain two embedding vectors $F_1^*$ and $F_2^*$.

The contrastive loss function is defined as:
\begin{equation}
\text{Loss} = \frac{1}{2} Y \cdot D_c^2 + \frac{1}{2} (1-Y) \cdot \max(0, m - D_c)^2
\end{equation}
where $D_c$ is the cosine distance:
\begin{equation}
D_c = 1 - \frac{F_1^* \cdot F_2^*}{\|F_1^*\| \|F_2^*\|} = 1 - \frac{\sum_{i=0}^{49} f_{1,i}^* f_{2,i}^*}{\sqrt{\sum_{i=0}^{49} (f_{1,i}^*)^2} \sqrt{\sum_{i=0}^{49} (f_{2,i}^*)^2}}
\end{equation}
$Y$ is a binary label ($Y=1$ for similar poses, $Y=0$ for dissimilar), and $m$ is a margin hyperparameter. We set $m=1.35$ based on experimental tuning.

\subsection{Pose Similarity Scoring}

Once trained, the model computes similarity scores between poses. Given two input images, we extract their embedding vectors and calculate cosine distance $D_c$. To provide an intuitive output, we map the distance $x=D_c$ to a score between 0 and 100 using a Gaussian-like function:
\begin{equation}
f(x) = \sigma \cdot e^{-\frac{1}{2}\left(\frac{x}{u}\right)^2}
\end{equation}
where $\sigma=100$ and $u=0.3$. When poses are identical, $x=0$ and the score is 100. As pose difference increases, the score decreases non-linearly.

\section{Experimental Evaluation}

\subsection{Datasets}

We evaluate on two widely used AQA benchmark datasets:

\textbf{AQA-7 Dataset:} AQA-7 comprises 1,189 video clips from seven different sports \cite{dfm_arxiv}. Each video has an expert quality score.

\textbf{FineDiving Dataset:} FineDiving \cite{finediving_paper} contains 3,000 diving videos with fine-grained annotations including action type, difficulty, temporal boundaries, and official scores.

\textbf{Training Pair Generation:} We generate positive pairs from high-scoring videos (above median) and negative pairs from different action categories or videos with large score differences.

\subsection{Evaluation Metric}

The standard metric is Spearman's Rank Correlation Coefficient ($\rho$) \cite{spearman_wiki}. This measures the consistency between predicted and expert rankings, with values closer to 1 indicating better performance.

\subsection{Implementation Details}

\textbf{HPE Module:} YOLOv5 and HRNet were pre-trained on the COCO dataset \cite{hrnet_github}.

\textbf{GCN-PSN Training:} Implemented in PyTorch with Adam optimizer, learning rate $1 \times 10^{-4}$, batch size 64, trained for 50 epochs.

\textbf{Baseline Methods:}
\begin{itemize}
\item \textbf{MLP-Baseline:} Removes the GCN module, directly feeds flattened coordinates into MLP.
\item \textbf{CoRe (I3D):} Video-based contrastive regression method using I3D features.
\item \textbf{TPT:} State-of-the-art Transformer-based method.
\end{itemize}

\subsection{Performance Comparison}

Table \ref{tab:performance} compares our method with baselines on AQA-7 and FineDiving datasets using Spearman's Rank Correlation $\rho$.
\begin{table}[htbp]
\caption{Performance Comparison on AQA-7 and FineDiving}
\begin{center}
\begin{tabular}{|l|c|c|}
\hline
\textbf{Method} & \textbf{AQA-7} & \textbf{FineDiving} \\
\hline
MLP-Baseline & 0.765 & 0.812 \\
CoRe (I3D) & 0.840 & 0.903 \\
TPT & 0.887 & 0.931 \\
\textbf{GCN-PSN (Ours)} & \textbf{0.851} & \textbf{0.915} \\
\hline
\end{tabular}
\label{tab:performance}
\end{center}
\end{table}

Key observations:
\begin{itemize}
\item GCN-PSN significantly outperforms MLP-Baseline, demonstrating the advantage of modeling skeletal topology.
\item GCN-PSN outperforms CoRe (I3D), indicating that topology-aware pose modeling can match or exceed video-level methods.
\item There remains a gap with TPT, as expected since TPT captures temporal dynamics while our method focuses on static poses.
\end{itemize}

\subsection{Ablation Study}

Table \ref{tab:ablation} shows the ablation study isolating the GCN contribution.
\begin{table}[htbp]
\caption{Ablation Study on AQA-7 Dataset}
\begin{center}
\begin{tabular}{|l|c|}
\hline
\textbf{Model Variant} & \textbf{Spearman's $\rho$} \\
\hline
MLP-Baseline (No GCN) & 0.765 \\
\textbf{GCN-PSN (Full Model)} & \textbf{0.851} \\
\hline
\end{tabular}
\label{tab:ablation}
\end{center}
\end{table}

The 8.6 percentage point improvement clearly demonstrates the indispensability of the GCN module for learning skeletal topology.

\subsection{Qualitative Analysis}

\begin{figure}[t]
  \centering
  \includegraphics[width=\columnwidth]{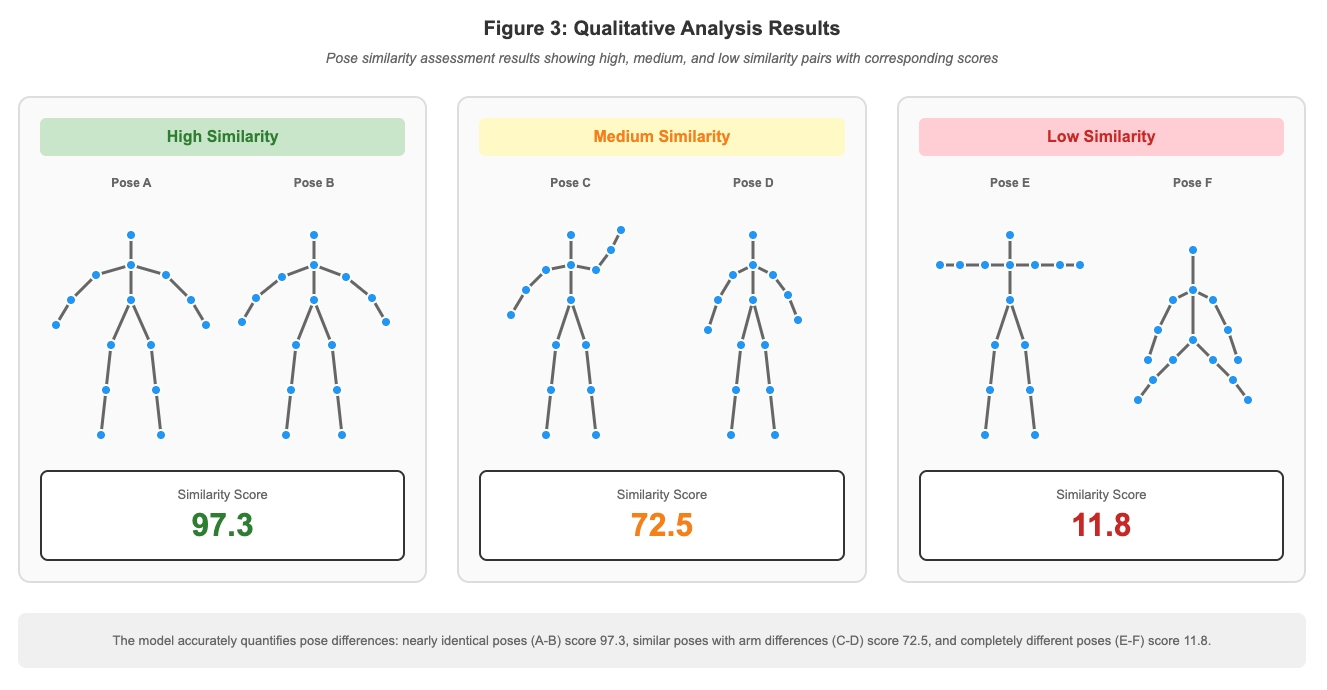}
  \caption{Qualitative analysis results. Left: High similarity (score: 97.3). Center: Medium similarity (score: 72.5). Right: Low similarity (score: 11.8). The model accurately quantifies subtle differences between poses.}
  \label{fig:qualitative}
\end{figure}

Fig.~\ref{fig:qualitative} shows qualitative results on pose pairs with varying similarity levels.

The results show that our model can distinguish between different actions at a macro level and is sensitive to subtle postural variations, quantifying them into scores that align with human perception.

\section{Conclusion and Future Work}

\subsection{Summary}

This paper proposes GCN-PSN, a novel framework based on Graph Convolutional Networks for pose similarity assessment in Action Quality Assessment. By explicitly modeling the human skeleton as a graph and leveraging GCNs to learn topological priors, we generate more discriminative and semantically meaningful pose embeddings than traditional coordinate vector methods. Comprehensive experiments on AQA-7 and FineDiving datasets demonstrate that GCN-PSN significantly outperforms topology-agnostic baselines and achieves performance comparable to state-of-the-art video-level methods.

\subsection{Limitations}

The current work has limitations that provide directions for future research:
\begin{itemize}
\item \textbf{2D Pose Dependence:} The model uses 2D keypoints, which suffer from depth ambiguity.
\item \textbf{Static Pose Focus:} The model evaluates single static poses and cannot assess dynamic motion quality.
\end{itemize}

\subsection{Future Directions}

Promising future directions include:
\begin{itemize}
\item \textbf{Extension to 3D Poses:} Applying GCNs to 3D skeleton data would eliminate 2D projection ambiguities \cite{3d_survey,egocentric_3d}.
\item \textbf{Spatio-Temporal Models:} Combining GCN-PSN with temporal models (LSTM/Transformer) would enable comprehensive static and dynamic assessment.
\item \textbf{Broader Applications:} The method can be applied to physical therapy, human-computer interaction, and ergonomic assessment in industrial settings.
\end{itemize}

In summary, by introducing topology-aware Graph Convolutional Networks, this paper provides an effective new paradigm for pose similarity assessment, laying a foundation for more refined action quality assessment systems.

\end{document}